\documentclass[conference]{IEEEtran}
\IEEEoverridecommandlockouts
\usepackage{cite}
\usepackage{amsmath,amssymb,amsfonts}
\usepackage{algorithmic}
\usepackage{graphicx}
\usepackage{textcomp}
\usepackage{xcolor}
\def\BibTeX{{\rm B\kern-.05em{\sc i\kern-.025em b}\kern-.08em
    T\kern-.1667em\lower.7ex\hbox{E}\kern-.125emX}} 
\begin{document}

\title{On the Resilience of Deep Learning for Reduced-voltage FPGAs
}

\author{ \small Kamyar Givaki$^{1}$, Behzad Salami$^{2}$, Reza Hojabr$^{1}$,     S. M. Reza Tayaranian$^{1}$, Ahmad Khonsari$^{1,3}$, Dara Rahmati$^{3}$, Saeid Gorgin$^{4}$,\\ Adrian Cristal$^{2,5,6}$, Osman S. Unsal$^{2}$. \\ 
$^{1}$ School of Electrical and Computer Engineering, University of Tehran, Tehran, Iran\\
$^{2}$ Barcelona Supercomputing Center, Barcelona, Spain\\ 
$^{3}$ School of Computer Sciences, Institute for Research in Fundamental Sciences (IPM), Tehran, Iran\\
$^{4}$ Iranian Research Organization for Science and Technology (IROST), Tehran, Iran\\
$^{5}$ Departament d’Arquitectura de Computadors, Universitat Politecnica de Catalunya, `
Barcelona, Spain\\
$^{6}$ Artificial Intelligence Research Institute (IIIA), Centro Superior de Investigaciones
Cient´ıficas (CSIC), Barcelona, Spain\\

givakik@ut.ac.ir, behzad.salami@bsc.es, r.hojabr@ut.ac.ir, m.taiaranian@ut.ac.ir, a\_khonsari@ut.ac.ir, dara.rahmati@ipm.ir,\\ gorgin@irost.ir, adrian.cristal@bsc.es, osman.unsal@bsc.es

 }

\maketitle

\begin{abstract}
 Deep Neural Networks (DNNs) are inherently computation-intensive and also power-hungry. Hardware accelerators such as Field Programmable Gate Arrays (FPGAs) are a promising solution that can satisfy these requirements for both embedded and High-Performance Computing (HPC) systems. In FPGAs, as well as CPUs and GPUs, aggressive voltage scaling below the nominal level is an effective technique for power dissipation minimization. Unfortunately, bit-flip faults start to appear as the voltage is scaled down closer to the transistor threshold due to timing issues, thus creating a resilience issue.

This paper experimentally evaluates the resilience of the training phase of DNNs in the presence of voltage underscaling related faults of FPGAs, especially in on-chip memories. \color{black} Toward this goal, we have experimentally evaluated the resilience of LeNet-5 and also a specially designed network for CIFAR-10 dataset with different activation functions of Rectified Linear Unit (Relu) and Hyperbolic Tangent (Tanh). We have found that modern FPGAs are robust enough in extremely low-voltage levels and that low-voltage related faults can be automatically masked within the training iterations, so there is no need for costly software- or hardware-oriented fault mitigation techniques like ECC. Approximately 10\% more training iterations are needed to fill the gap in the accuracy. This observation is the result of the relatively low rate of undervolting faults, i.e., \textless 0.1\%, measured on real FPGA fabrics. We have also increased the fault rate significantly for the LeNet-5 network by randomly generated fault injection campaigns and observed that the training accuracy starts to degrade. When the fault rate increases, the network with Tanh activation function outperforms the one with Relu in terms of accuracy, e.g., when the fault rate is 30\% the accuracy difference is 4.92\%.\color{black} 

\end{abstract}


\section{Introduction}
\color{black}

Hardware accelerators are designed to perform required computations in a specific application efficiently \cite{arcas2015hardware, salami2015hatch, salami2017axledb, salami2016accelerating, melikoglu2019novel, gizopoulos2019modern}. Deep Neural Networks (DNNs) need a huge amount of computations, which categorizes them as power- and energy-hungry applications. Using hardware accelerators is a promising solution to answer this requirement. Recently, many accelerators based on Graphics Processing Units (GPUs) \cite{nurvitadhi2017can}, Field Programmable Gate Arrays (FPGAs) \cite{acc1,acc2,acc5,review3}, and Application Specific Integrated Circuits (ASICs) \cite{reagen2016minerva,skippynn} have been proposed for various DNNs. Among them, FPGAs have unique features, which makes them increasingly popular, thanks to their massively parallel architecture, reconfiguration capability, data-flow execution model, and the recent advances on the High-Level Synthesis (HLS) tools. However, the power consumption of FPGAs is still a key concern, especially comparing against equivalent ASICs, and FPGAs can be at least an order of magnitude less energy-efficient than ASIC designs \cite{nurvitadhi2016accelerating}. To mitigate this gap, aggressive voltage underscaling is an effective solution \cite{salami2018comprehensive, salami2018aggressive}, by considering the quadratic saving in dynamic power and exponential saving in static power \cite{salamin2019selecting}; aggressive voltage underscaling can be described as decreasing the supply voltage of either whole or some components of a circuit below the nominal voltage which is set by the vendor. However, some reliability issues might appear as a result of the circuit delay increase. These timing issues may cause some faults; therefore, the circuit can produce the wrong results. Generally mitigating these effects is done by hardware design changes\cite{razor,understanding} or by using the built-in Error Correction Code (ECC) of FPGAs \cite{salami2019evaluating}. These types of efforts are also carried out under research projects like LEGaTO \cite{salami2019legato, cristal2018legato}. DNN applications are inherently tolerant to some faults. This is a unique property that distinguishes DNNs as good applications to apply aggressive voltage underscaling. The reason is that no additional hardware or technique would be needed to be applied to mitigate the effects of aggressive voltage underscaling in DNN applications.  

\begin{figure}[t]
\centerline{    \includegraphics[width=\linewidth]{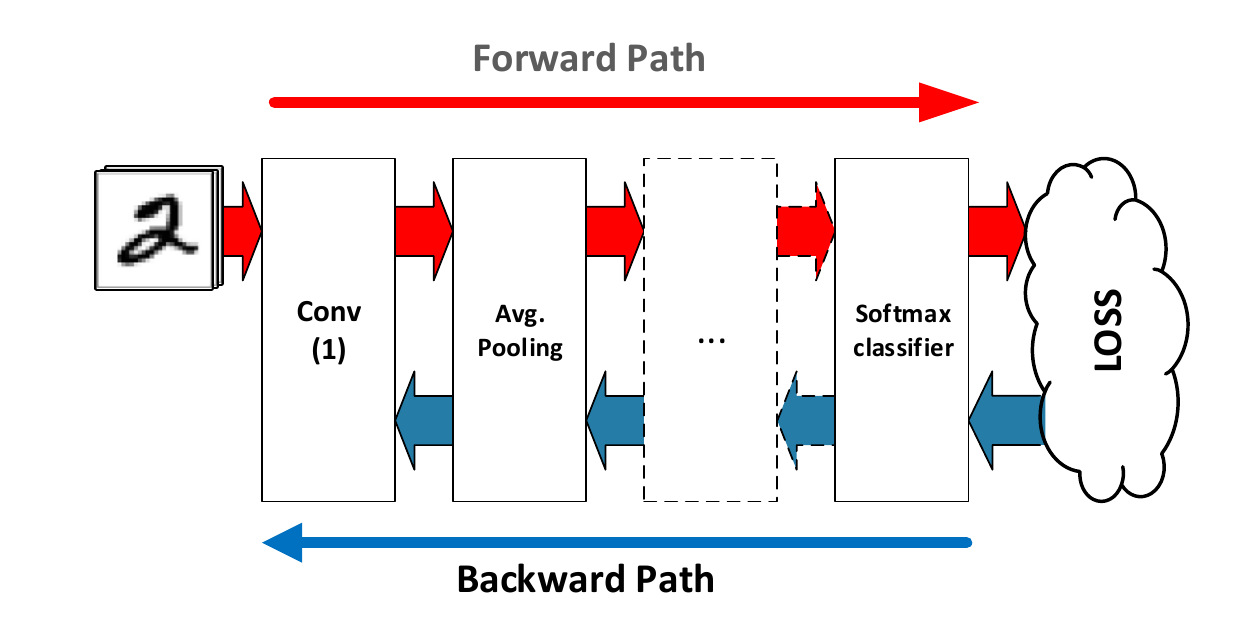}}
\caption{High-level abstraction of an iteration in the training process of LeNet-5 network.}
\label{fig:train}
\end{figure}

Typically, a working DNN has two phases of operation. The first phase, \textit{training}, is the process of tuning the parameters of a specific network. Training is an iterative task in which sample inputs are iteratively injected into the network. Then, the predicted results of the network are evaluated by the desired results using a specific function known as \textit{loss function}. After that, the loss is propagated backward for parameter tuning. This process continues until the loss of the network decreases under a threshold value. In most cases, the process of training a network is performed only
once. Fig.~\ref{fig:train} shows a high-level abstraction of one iteration in the training process in which a sample enters the network, then the loss of the network is computed and the loss propagates to the first layer through the backward path. In the second phase (known as  \textit{inference}), a sample is injected into the network, and the network generates an output based on the parameters that are learned in the training process. In this paper, our focus is on the training phase of neural networks as it is more energy-hungry than the inference phase, and reducing the power consumption by voltage underscaling can be directly translated to lower energy consumption.\color{black}
\color{black}
The effect of the faults is mainly investigated in the inference phase of DNNs in hardware, software, embedded, and HPC platforms \cite{kumar2017survey}. There are some recent efforts on the training phase too \cite{zhang2018analyzing, hacene2019training}, but these faults are related to manufacturing defects or soft errors.
 However, to the best of our knowledge, no work addresses undervolting related faults of COTS hardware in the training phase of DNNs. 

In this paper, we contribute by examining the resilience of DNN training by using the on-chip SRAM-based memory fault maps of real FPGA fabrics \cite{salami2018comprehensive}. Note that SRAMs play an important role in the structure of DNN accelerators \cite{guo2017survey}. They also have a significant contribution in range of 30\%-70\% on the total power consumption of such DNN systems \cite{salamat2019workload, conti2018xnor}. Thus, the focus of this paper is on the on-chip memories. Our experiments confirm the idea that the faults related to the aggressive voltage underscaling are masked in the training process due to the inherent fault resiliency of DNNs. The fault rate of undervolted FPGAs is less than 0.1\%, and this fault rate has a negligible negative effect on the training process.

We found that with the higher fault rates of at least 25\%, the DNN accuracy can be affected by 6.25\%  (with the same number of iterations). This gap can be filled with more training iterations. It should be mentioned that the accuracy of the network with the Hyperbolic Tangent (Tanh) activation function has been less affected by increasing the fault rate. 
\color{black}

In a nutshell, we evaluate the resilience of DNN training in the presence of FPGA undervolting faults. More specifically, the contributions of this paper are listed below:
\begin{itemize}
    \item The DNN training is inherently robust for undervolting-related faults, evaluated on the fault maps of real FPGA fabrics that are publicly available. This observation is due to the relatively low fault rate for modern FPGAs that is measured up to 0.1\%. 
    \item We generate higher fault rates with uniform distribution to complete our experiments. For the LeNet-5 network, the fault rate of at least 25\% can significantly affect the DNN accuracy.
  \end{itemize} 

The rest of the paper is organized as follows. The experimental methodology is introduced in Section \ref{sec:methodology}. The obtained results are presented and discussed in Section \ref{sec:results}. We review the related work in Section \ref{sec:related} and finally the paper is concluded in Section \ref{sec:conclusion}.  

\section{Experimental Methodology}
\label{sec:methodology}
Our experiments are based on injecting undervolting-related faults into the inputs, weights, and intermediate values generated during the DNN training phase. To evaluate the resilience behavior, we compare the accuracy and loss of the faulty DNNs with the baseline without any faults. Below, we elaborate on the experimental methodology, including the fault and the DNN models, as well as the overall experimental setup. 

\subsection{Fault Model}

Voltage underscaling below the minimum safe voltage level, i.e., $V_{min}$, can result in timing faults. In \cite{salami2018comprehensive}, this technique is investigated for modern FPGAs, specifically on SRAM-based on-chip memories. Reference \cite{salami2018comprehensive} reports that the fault rate increases exponentially when the voltage is decreased below $V_{min}$. At the lowest voltage level that could be practically underscaled, i.e., $V_{crash}$, (almost half of the default voltage level, i.e., $V_{nominal}$), the maximum fault rate observed is less than 0.1\%. Also, it has been shown that the faults show a permanent behavior for a specific device, and their location does not typically change at a fixed voltage level. The undervolting fault maps, i.e., the distribution of faults in physical locations of memories at different voltage levels below the $V_{min}$, are released in \cite{fault-map} publicly. The undervolting fault maps are unique and per-FPGA, due to the process variation effects, demonstrated on VC707 and KC705 in \cite{salami2018comprehensive}. More specifically, it has been explored in \cite{salami2018comprehensive} that faults appear in [$V_{min}=0.6V$, $V_{crash}=0.54V$] and [$V_{min}=0.59V$, $V_{crash}=0.53V$] for a maximum of 23706 and 2274 faults for VC707 and KC705, respectively. It should be noted that in both FPGAs, the $V_{nominal}=1V$. We utilize the fault map from \cite{fault-map} for each FPGA that precisely illustrates the location of flipped bits in each FPGA under different underscaled voltages.

We use these publicly-available undervolting fault maps and inject them into the DNN training and monitor the accuracy. Note that the total size of the available FPGA on-chip memories is limited, \textit{e.g.,} 4.5 MB, and 1.9 MB for VC707 and KC705, respectively. We use this memory to store inputs of the network, weights of the network, and intermediate values of computations, \textit{i.e.,}, values of losses in each iteration during the training. So, memories are crucial components in implementing DNNs, and a significant fraction of the total power consumption of the whole system is related to these components. Therefore decreasing the power consumption in block RAMs can be directly translated to overall system power consumption reduction.

\subsection{DNN Models}

\color{black}

  \begin{figure}[!t]
\centerline{\includegraphics[width=\linewidth ]{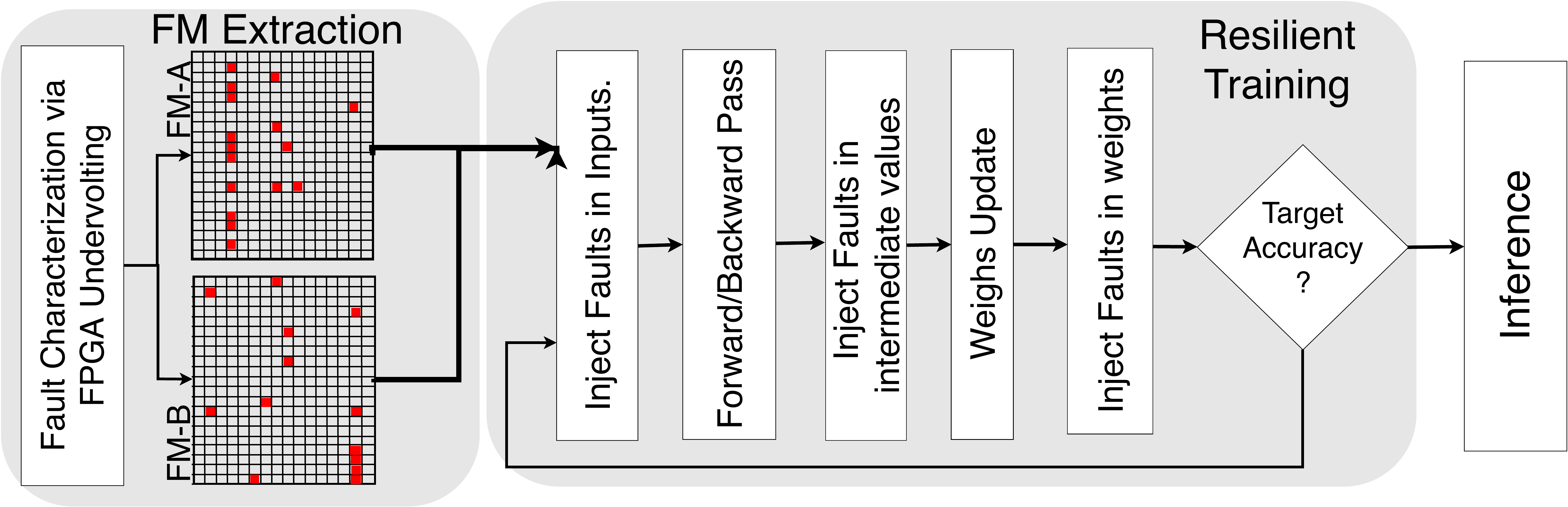}}
\caption{Overall methodology}
\label{fig:full}
\end{figure}

 To evaluate the impact of voltage underscaling on the training of neural networks, we apply our model on two convolutional neural networks. The first DNN model is LeNet-5\cite{lenet}. As illustrated in Table~\ref{table:lenet}, the network has two convolutional layers, each of them followed by an average pooling (sub-sampling) layer. Then two fully-connected layers and a softmax layer are placed to generate the desired output. We use the MNIST dataset \cite{mnist} to train this network. MNIST contains 60000 samples of handwritten digits to train the network and also includes 10000 samples to test the training process. each sample is a $28 \times 28 $ gray-scale image. We train this network for classification MNIST by more than $10K$ iterations, and the top-1 classification accuracy achieved is 98.6\% (normally, the classification accuracy reaches 99.5\%, but more training iterations are needed). \color{black}

The second network is a special architecture that is designed to classify images of the CIFAR-10 dataset \cite{cifar10}, and the details of this network are shown in Table~\ref{table:cifararch}. The dataset contains 50000 training images with a size of $32 \times 32$ pixels and also 10000 test images. After more than $100K$ of training iterations, the top-1 classification accuracy was 85.7\% (the reported accuracy is for the case with Rectified Linear Unit (Relu) Activation function).\color{black} It is constructed from four convolutional layers, one sigmoid fully connected layer, and a softmax layer. 
\color{black}

 \begin{table}[t]
 \centering
\caption{The detailed architecture of LeNet-5. \cite{lenet}}.
\label{table:lenet}
\begin{tabular}{|c|c|c|c|c|}
\hline
\begin{tabular}[c]{@{}c@{}}Layer\\ type\end{tabular} & Kernel size  & Stride & \begin{tabular}[c]{@{}c@{}}\# of \\ channels\end{tabular} & \begin{tabular}[c]{@{}c@{}}Activation \\ function\end{tabular} \\ \hline \hline
Conv                                                 & $5 \times 5$ & 1      & 6                                                         & Relu or Tanh                                                           \\ \hline
Avg. pooling                                         & $2 \times 2$ & 2      & --                                                         & --                                                              \\ \hline
Conv                                                 & $5 \times 5$ & 1      & 16                                                        & Relu or Tanh                                                           \\ \hline
Avg. pooling                                         & $2 \times 2$ & 2      & --                                                         & --                                                              \\ \hline
FC layer                                             & --           & --      & --                                                         & Sigmoid                                                        \\ \hline
FC layer                                             & --            & --      & --                                                         & Sigmoid                                                        \\ \hline
Softmax classifier                                   & --            & --      & --                                                         & Softmax                                                        \\ \hline
\end{tabular}

\end{table}

\begin{table}[t]
\centering
\caption{Detailed architecture of the CNN which is used for CIFAR-10 dataset.}
\label{table:cifararch}
\begin{tabular}{|c|c|c|c|c|}
\hline
\begin{tabular}[c]{@{}c@{}}Layer \\ type\end{tabular} & \begin{tabular}[c]{@{}c@{}}Kernel \\ size\end{tabular} & Stride & \begin{tabular}[c]{@{}c@{}}\# of \\ Channels\end{tabular} & \begin{tabular}[c]{@{}c@{}}Activation\\ function\end{tabular} \\ \hline \hline
Conv                                                  & $3 \times 3$                                           & 1      & 32                                                        & Relu or Tanh                                                          \\ \hline
Conv                                                  & $3 \times 3$                                           & 1      & 32                                                        & Relu or Tanh                                                          \\ \hline
Max pooling                                           & $2 \times 2$                                           & --     & --                                                        & --                                                            \\ \hline
0.3 Dropout                                           & --                                                     & --     & --                                                        & --                                                            \\ \hline
Conv                                                  & $3 \times 3$                                           & 1      & 64                                                        & Relu or Tanh                                                          \\ \hline
Conv                                                  & $3 \times 3$                                           & 1      & 64                                                        & Relu or Tanh                                                       \\ \hline
Max pooling                                           & $2 \times 2$                                           & --     & --                                                        & --                                                            \\ \hline
0.4 Dropout                                           & --                                                     & --      & --                                                        & --                                                            \\ \hline
FC layer                                              & --                                                     & --     & --                                                        & Sigmoid                                                       \\ \hline
FC layer                                              & --                                                     & --     & --                                                        & Softmax                                    
\\ \hline
\end{tabular}
\end{table}

\color{black}

\subsection{Overall Methodology}
The overall experimental methodology is shown in Fig.~\ref{fig:full}. As seen, we utilize the fault map of the FPGA memories for different chips and inject these faults into the inputs (pixels of input images), weights of the DNN, and all values which are generated in both forward and backward paths of training process based on the location of them on block RAMs. After updating the faulty weights in each iteration, we repeat the process for other iterations, where faults appear at the same location, due to the permanent behavior of the undervolting faults.

In our simulations, after iteration $i$, the updated weights are obtained by injecting faults to the values based on the position that values are stored in block RAMs. Weights that are obtained in iteration $i$ are used for the next iteration of training. So in $(i+1)-th$ iteration, the training process of network tries to eliminate the effects of faults in previous iterations of the training. 
 It should be mentioned that the training for all the voltages has been performed in the same iterations, and the loss function in all experiments is categorical cross-entropy. 
 
\begin{figure}[!t]
\centerline{    \includegraphics[height=8cm,width=0.35\textwidth]{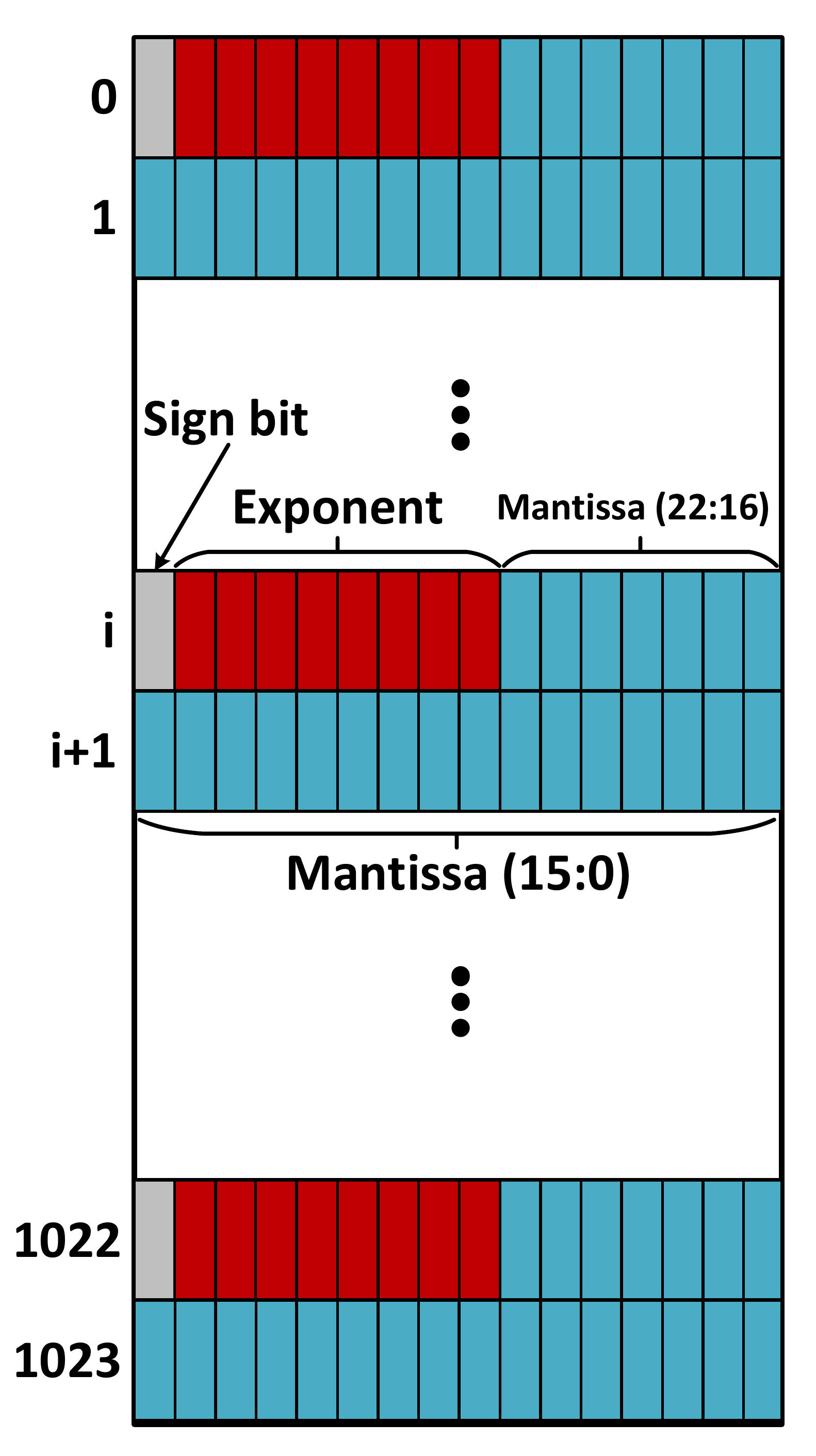}}
\caption{Storing a 32-bit floating-point number (input, weight, and intermediate data) to two rows of a single block ram with the size of 16K bit, i.e., a matrix of 1024 rows and 16 columns. The intersection of each row and column represents a bitcell.}
\label{fig:fpconversion}
\end{figure}

  We use single-precision floating-point numbers (32-bits) to represent inputs, weights, and intermediate variables; unlike inference, the training process typically requires floating-point computation. Each word of block RAMs in employed FPGAs has 16-bits. So, two words of the block RAM are needed to store a single-precision floating-point number.
 A floating-point number is stored in two words of block RAM in a way that is shown in Fig.~\ref{fig:fpconversion}. \color{black} Hence, according to the location of reported faults in block RAMs, some bit-flips may occur in sign-bit, mantissa, or exponent of the floating-point number.  
 \color{black}

 \begin{figure*}[t]
 \centering
\centerline{    \includegraphics[width=\linewidth]{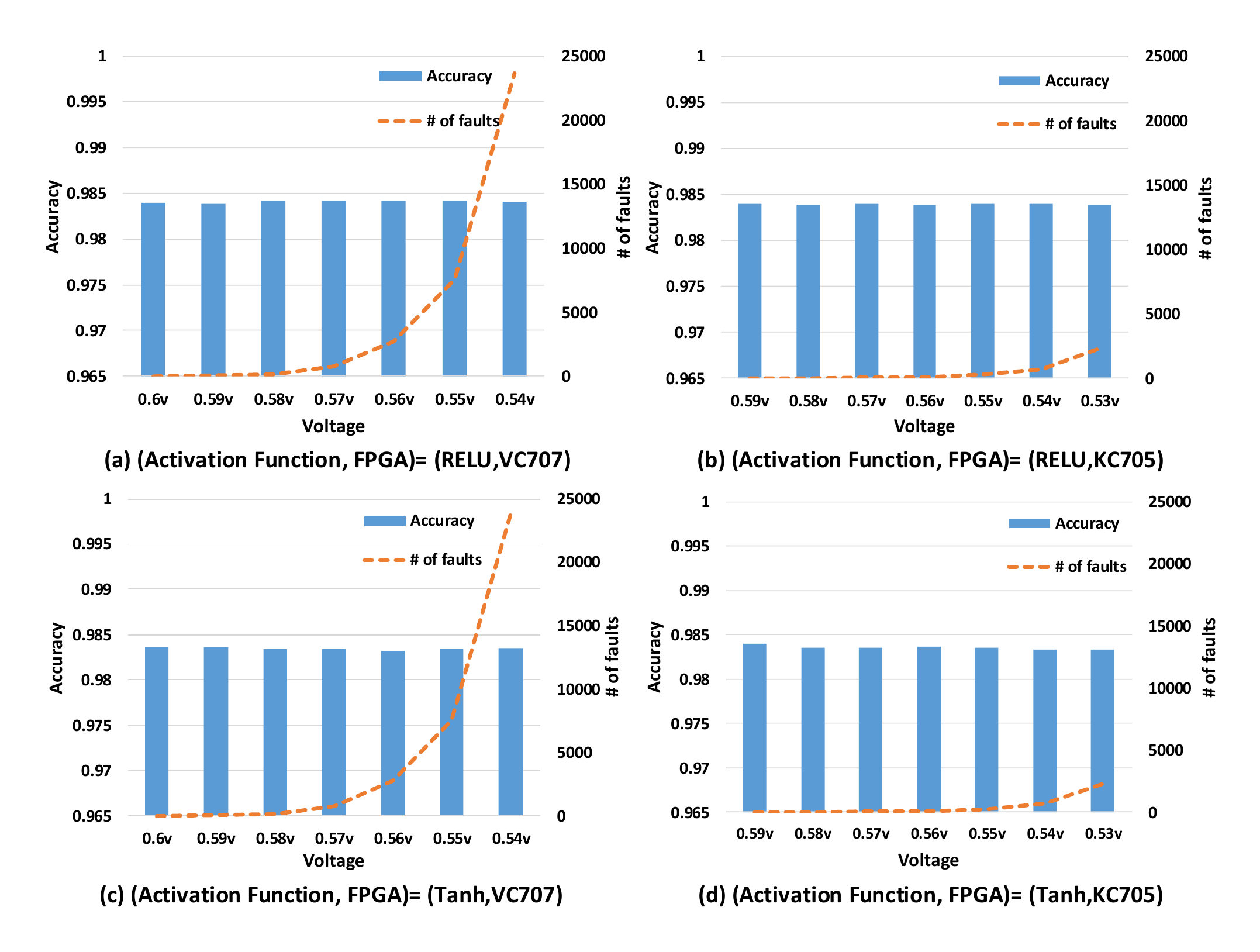}}
\caption{The training accuracy of the LeNet-5 network in the classification of MNIST dataset under different voltages and with different number of faults  for (Activation function, FPGA).}
\label{fig:mnistacc}
\end{figure*}
  Values of input images, weights, and intermediate values are stored in block RAM in the following manner: first, several block RAMs (two block RAMs for MNIST and six for CIFAR-10) of the FPGA are assigned to store values of pixels of the input images. These required block RAMs are selected randomly. According to the architecture of the network, several block RAMs are reserved for storing the latest value of weights. The two employed networks are small enough for all of their weights can be stored in block RAMs offered by FPGAs. These block RAMs are also selected randomly.
  Intermediate values that are generated in the process of training are written to other parts of block RAMs.
  When the capacity of \color{black} the FPGA's block RAMs has \color{black} been reached, the new intermediate values are substituted by previously generated ones. The replacement policy is First-In-First-Out (FIFO), in which the oldest intermediate value has replaced with the latest generated intermediate value. 
 
  By writing all values in the above-mentioned order, it is possible to determine the exact position of each variable (Input, weight, intermediate value). So, it can be possible to simulate the impact of voltage underscaling on the values and also the whole network, using the fault map of FPGA presented in \cite{fault-map}. 
  \color{black}
\color{black}

\color{black}
\section{Experimental Results}
\label{sec:results}

\begin{figure*}[h]
 \centering
\centerline{    \includegraphics[width=\linewidth]{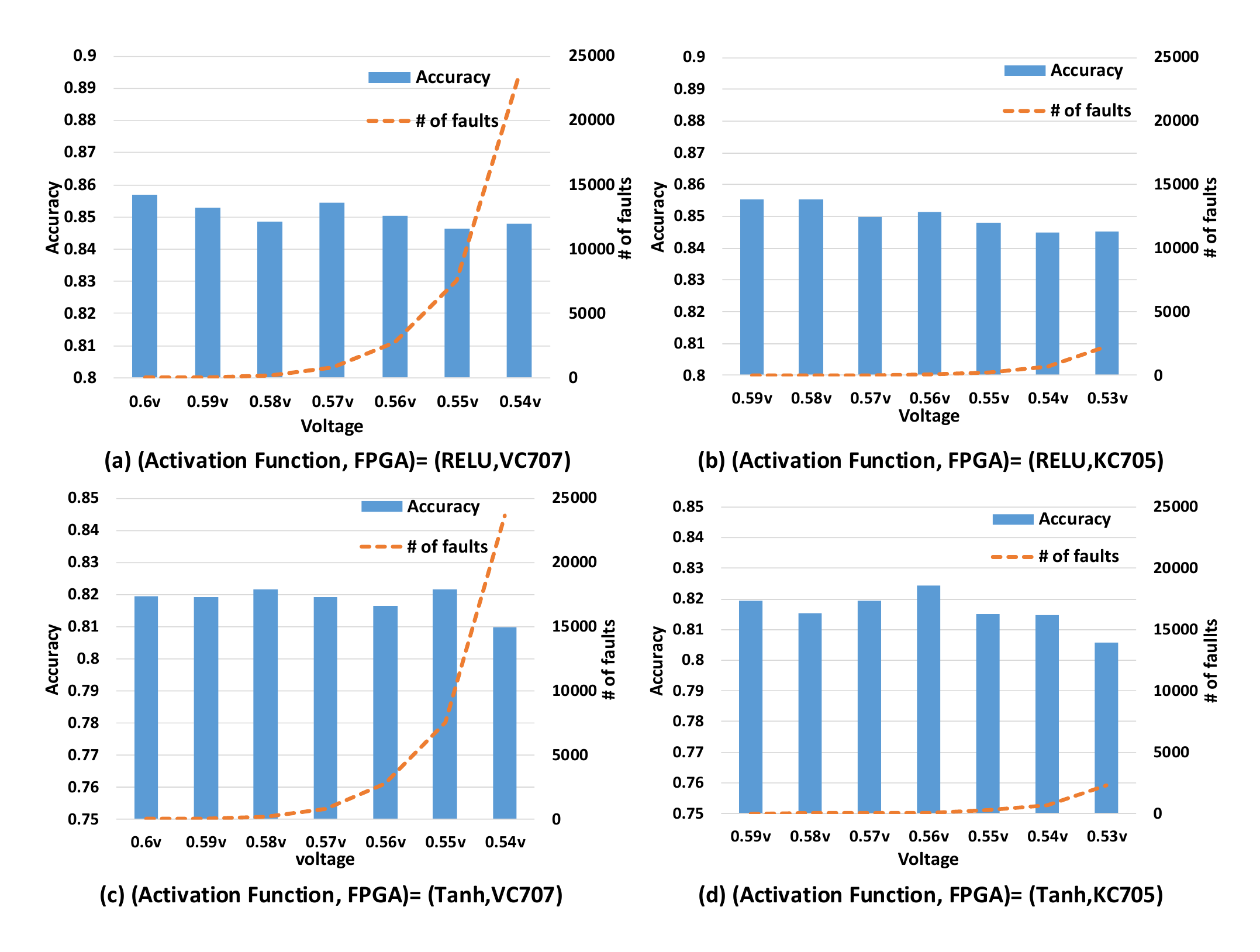}}
\caption{The training accuracy of the network that is used in the classification of the CIFAR-10 dataset under different voltages and with the different number of faults for (Activation function, FPGA). }
\label{fig:cifaracc}
\end{figure*}
Fig.~\ref{fig:mnistacc} shows the Lenet-5 network accuracy when it is used for the classification of the MNIST dataset. Fig.~\ref{fig:mnistacc} (a) illustrates the accuracy of the network, which is simulated using the VC707 fault map, and the activation function is Relu. As can be seen in this figure, decreasing the voltage results in an increase in the fault rate, which results in a minimal and negligible decrease in the network accuracy. As illustrated in Fig.~\ref{fig:mnistacc} (b), the accuracy of the network, which is simulated using the VC705 fault map and Relu activation function decreases slightly. Fig.~\ref{fig:mnistacc} (c) and (d) show the simulations when the activation functions are Tanh. As Fig.~\ref{fig:mnistacc} shows, when the real fault maps are used for simulations, there is not much difference in the accuracy results when the activation functions for convolutional layers are either Relu or Tanh. \color{black}

  \begin{figure}[h]
 \centering
\centerline{    \includegraphics[width=\linewidth]{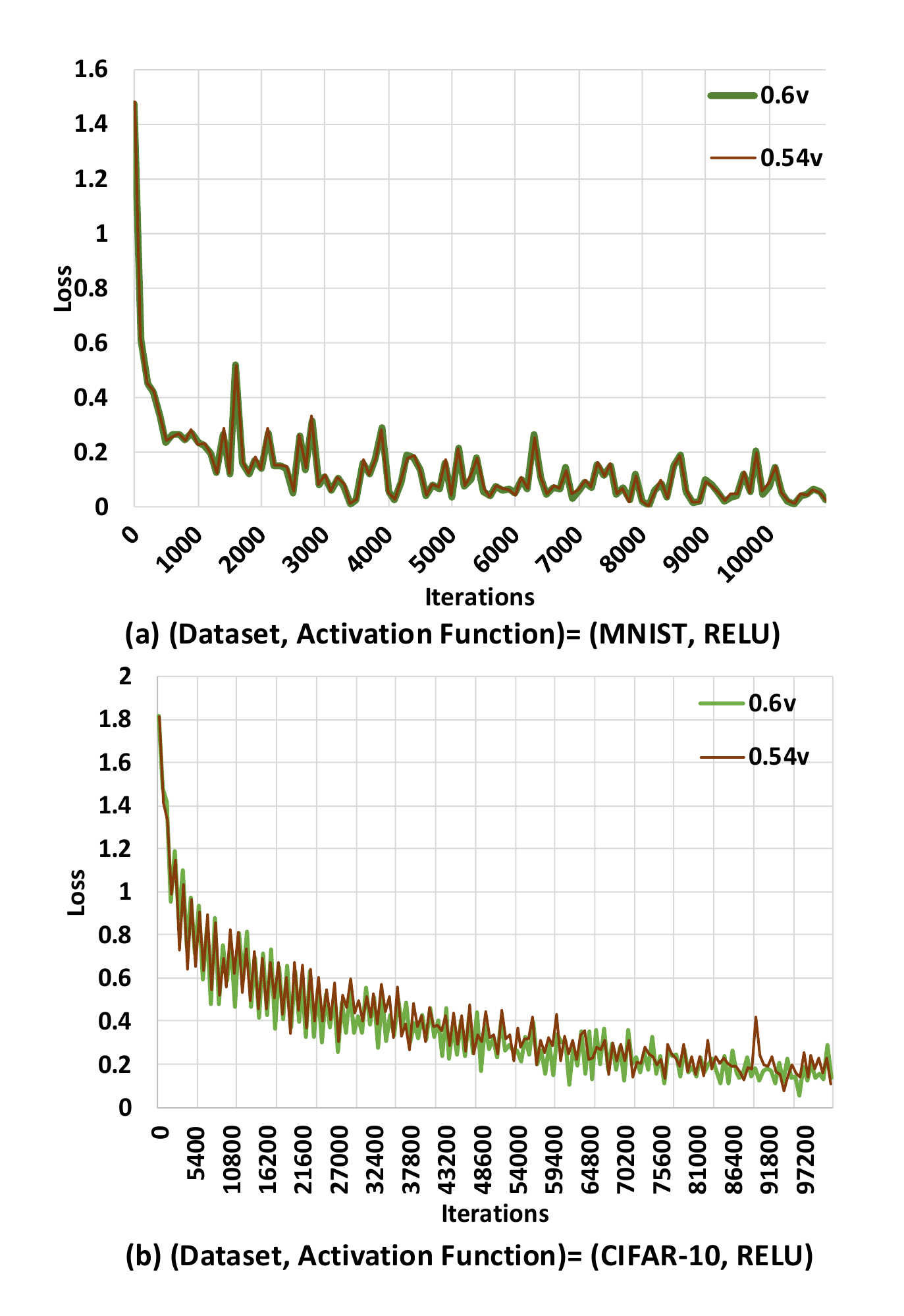}}
\caption{The Loss of the networks for training using VC707 fault map (a) MNIST (b) CIFAR-10.}
\label{fig:Loss}
\end{figure}

Fig.~\ref{fig:cifaracc} illustrates the accuracy of the classification of the CIFAR-10 dataset. We use the network that its structure is shown in Table~\ref{table:cifararch}. \color{black} When the network is simulated by the VC707 fault map, and activation function is Relu, as shown in Fig.~\ref{fig:cifaracc} (a), the accuracy of the network is negligibly decreased by supply voltage decrease. Fig.~\ref{fig:cifaracc} (b) shows a similar trend for VC705 fault map and Relu activation function. The accuracy of the network is reduced when we substitute the Relu activation function by Tanh. Although, as depicted in Fig.~\ref{fig:cifaracc} (c) and (d), the trend of accuracy changes is similar to the trend of the Relu activation function in which the accuracy decreases with voltage decrease. In this case, the network with the Relu activation function is more accurate. It should be mentioned that the training for all the voltages has been performed with the same number of iterations (more than $10K$ iterations for MNIST and more than $100K$ for the CIFAR-10). 

Fig.~\ref{fig:Loss} shows the loss values for the training of these networks for two voltages ($V_{min}$ and $V_{crash}$). Fig.~\ref{fig:Loss} (a) illustrates the loss of the LeNet-5 network when VC707 is employed. Since the LeNet-5 network is a small network with a low number of parameters, the loss for both the voltages follows the same trend, and there is no significant difference between the loss values.

On the other hand, Fig.~\ref{fig:Loss} (b) shows as the network parameters increase, the gap between the loss values for two different voltages reveals (the network that is used for classification of the CIFAR-10 has more parameters than LeNet-5). This gap can be interpreted as follows: The lower voltage can decrease the convergence rate of the network. In other words, if we decrease the voltage, the training process needs more iterations typically to reach a specific accuracy point. For example, Table~\ref{t:iter} shows the difference between the iterations to reach the accuracy of 98\% in classification of MNIST in several cases and the iterations to reach the accuracy of 80\% for CIFAR-10.
 Table~\ref{t:iter} reveals, on average, 10\% additional iterations can handle the effect of these faults in the accuracy of the networks. For example, to reach 98\% top-1 accuracy in MNIST, approximately 200 more iterations are required.
 
 Reference \cite{salami2018comprehensive} has shown that in a VC707 FPGA, by decreasing the supply voltage of block RAMs from $V_{min}$ to $V_{crash}$, the power consumption of block RAMs is decreased by 40\%; however, this reduction can lead to reliability issues and result in some faults in the content of block RAMs. Our observations show that these faults can be masked in the process of training. As previously mentioned, in modern DNNs, 30\%-70\% of the total power consumption of the system is related to SRAMs. By combining the two above mentioned facts, it can be inferred that aggressive voltage underscaling can decrease the total power consumption of the system by at least 10\%.
\color{black}

\begin{table}[!t]
\centering
\caption{Comparison of the number of iterations to reach to a specific accuracy point. }
\label{t:iter}
\begin{tabular}{|c|c|c|c|c|}
\hline
Dataset                                                                    & \begin{tabular}[c]{@{}c@{}}Reference \\ accuracy\end{tabular} & Device                 & Voltage & Iterations \\ \hline   \hline
{\begin{tabular}[c]{@{}c@{}}MNIST\\ (Relu)\end{tabular}}    & {98\%}                                        & {VC707} & $V_{min}$=0.6V & 4950       \\ \cline{4-5} 
                                                                           &                                                              &                        & $V_{crash}$=0.54V   & 5200       \\ \cline{3-5} 
                                                                           &                                                              & {KC705} & $V_{min}$=0.59V & 4900       \\ \cline{4-5} 
                                                                           &                                                              &                        & $V_{crash}$=0.53V   & 5050       \\ \hline
{\begin{tabular}[c]{@{}c@{}}CIFAR-10\\ (Tanh)\end{tabular}} & {80\%}                                        & {VC707} & $V_{min}$=0.6V & 47800      \\ \cline{4-5} 
                                                                           &                                                              &                        & $V_{crash}$=0.54V   & 51200      \\ \cline{3-5} 
                                                                           &                                                              & {KC705} & $V_{min}$=0.59V & 43800      \\ \cline{4-5} 
                                                                           &                                                              &                        & $V_{crash}$=0.53V   & 61000       \\ \hline
\end{tabular}
\end{table}

 
\color{black}

The reported fault rate in real FPGA fault maps is under 0.1\%. To investigate the resilience of the training process to higher fault rates, we perform an experiment in which we generate fault maps with higher fault rates. \color{black} Faults are randomly distributed in the whole block RAM spaces with uniform distribution, and simulations are performed for VC707 FPGA block RAM size and Lenet-5 network. \color{black}
 \begin{figure}[h]
 \centering
\centerline{    \includegraphics[width=\linewidth]{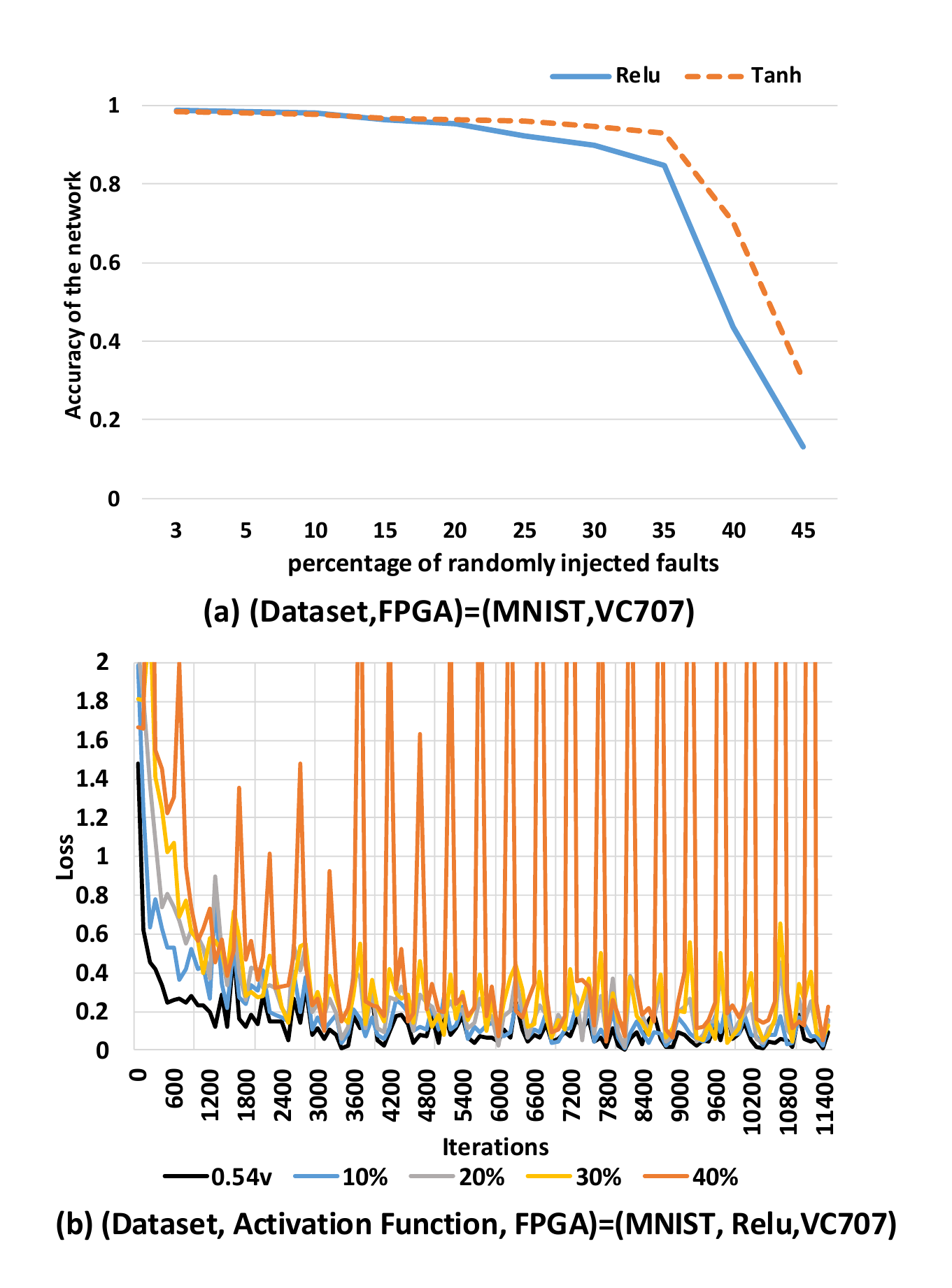}}
\caption{ (a) The accuracy comparison of Relu and Tanh activation functions for randomly generated fault maps (b) Comparison of the network loss for several random generated fault maps with  0.54V real fault map for VC707 and LeNet-5 when the activation function is Relu.}
\label{fig:rand}
\end{figure}

 Fig.~\ref{fig:rand} (a) shows the training accuracy of LeNet-5 in two cases: In the first case, the activation functions of convolutional layers are Relu, and in the second one, the activation functions of convolutional layers are Tanh. In both cases, the accuracy remains high when the injected fault rate is lower than 25\%. Then the accuracy decreases, and at a point between 30 and 40\%, the accuracy curve breaks. Fig.~\ref{fig:rand} (b) shows the loss values for some fault rates. \color{black} As seen, the network with the Tanh activation function outperforms the one with Relu since the injected fault rate is more than 15\%; the Tanh network has 1.09\% better accuracy than the Relu one. The gap between these two curves extends to 4.92\% when the injected fault rate increases to 30\%. It can be inferred that in situations that fault rate is high, using the Tanh activation function may be helpful.

\color{black}

\section{Related Works}
\label{sec:related}

With technology scale developing, the resilience of DNNs can be significantly affected due to the fabrication process uncertainties, soft-errors, harsh and noisy environments, aggressively low-voltage operations, among others. Hence, recently, the resilience of DNNs has been studied in different abstraction levels. A vast majority of the previous works in this area belong to the DNN inference phase, including simulation-based efforts \cite{zhang2019fault, li2017understanding, choi2019sensitivity, salami2018resilience} and works on the real hardware \cite{zhang2018thundervolt, reagen2016minerva, pandey2019greentpu, chandramoorthy2019resilient}. The verification of the simulation-based works on the real fabric can be a crucial concern; also, the real hardware works are mostly performed on the customized ASICs, which of course, reproducing those results on the COTS systems is a crucial question. 

On the other hand, there are not thorough efforts on the resilience of the DNN training phase; recent works in part cover the study in this area \cite{review1, kim2018energy, kim2018matic, zhang2018analyzing, hacene2019training}. For instance, \cite{kim2018energy, kim2018matic} have analyzed only the fully-connected model of DNNs, \cite{zhang2018analyzing} carried out the analysis on a customized ASIC model of the DNN, and finally, \cite{hacene2019training} performed a simulation-based study. Our paper extends the study on the resilience of the DNN training, especially by using the fault maps of low-voltage SRAM-based on-chip memories of real FPGA fabrics.    

Our experimental methodology is based on emulating the real fault maps of FPGA-based SRAM memories during the DNN training iterations. A similar approach has been considered for real DRAMs, as well~\cite{koppula2019eden, review2}. Unlike the fully software-based approaches, our study is based on real fault maps, which can lead to more precise study. Also, unlike the fully hardware-based approach, our study is more facilitated and can be easily expanded for many different applications. In other words, our approach has the advantage of both full software \cite{chang2019assessing}, and fully real hardware \cite{bertran2014voltage} resilience study approaches, similar to recent works \cite{denkinger2019impact, koppula2019eden, chatzidimitriou2019assessing}.

\section{Conclusion}
\label{sec:conclusion}
In this paper, we experimentally evaluate the effect of aggressive voltage underscaling of FPGA block RAMs on the training phase of deep neural networks. Simulation results show that the training process of deep neural networks is resilient to faults that are generated because of the reduced-voltage supply. We observed that due to the low fault rate of real FPGA fabrics of up to 0.1\%, the effect of these faults on the accuracy of the network is negligible and can be compensated, on average,  by 10\% more iterations in training.

Furthermore, the training process is resilient to fault rate more than the fault rate of real FPGAs. Additionally, our simulations show that with injecting 25\% random faults to memory, the accuracy of the LeNet-5 Network in the classification of the MNIST dataset is only 6.25\%  for Relu activation function, and 2.75\% for Tanh activation function lower than the training with no faults (in the same number of iterations). As an ongoing work, we are going to repeat our model on real FPGAs.

\section{Acknowledgments}

The research leading to these results has received funding
from the European Union’s Horizon 2020 Programme under
the LEGaTO Project (www.legato-project.eu), grant
agreement n$^\circ$  780681.

\bibliographystyle{unsrt}
\bibliography{conference_041818.bib}
\end{document}